\documentclass[10pt,twocolumn,letterpaper]{article}

\usepackage[pagenumbers]{styles/wacv} 

\usepackage{graphicx}
\usepackage{amsmath}
\usepackage{amssymb}
\usepackage{booktabs}

\usepackage[pagebackref,breaklinks,colorlinks]{hyperref}

\usepackage[capitalize]{cleveref}
\crefname{section}{Sec.}{Secs.}
\Crefname{section}{Section}{Sections}
\Crefname{table}{Table}{Tables}
\crefname{table}{Tab.}{Tabs.}

\usepackage[accsupp]{axessibility} 

\usepackage[dvipsnames]{xcolor}
\usepackage{tikz}
\usepackage{microtype}
\usepackage{algorithm}
\usepackage{algpseudocode}

\DeclareRobustCommand*\circled[1]{\tikz[baseline=(char.base)]{
            \node[shape=circle,draw,inner sep=1pt] (char) {\textbf{#1}};}}
            
\DeclareMathOperator*{\argmin}{arg\,min}

\def\wacvPaperID{228} 
\def\confName{WACV}
\def\confYear{2025}

\begin{document}

\title{Pre-trained Multiple Latent Variable Generative Models are good defenders against Adversarial Attacks}

\author{Dario Serez$^{1,3}$ \hspace{5pt} Marco Cristani$^2$ \hspace{5pt} Alessio {Del Bue}$^1$ \hspace{5pt} Vittorio Murino$^{1,2,3}$ \hspace{5pt} Pietro Morerio$^1$\\
$^1$Istituto Italiano di Tecnologia, Italy \hspace{10pt} $^2$University of Verona, Italy  \hspace{10pt} $^3$University of Genoa, Italy\\
{\tt\small \{dario.serez, alessio.delbue, vittorio.murino, pietro.morerio\}@iit.it}\\
{\tt\small marco.cristani@univr.it}
}

\twocolumn[{%
\renewcommand\twocolumn[1][]{#1}%
\maketitle
\begin{center}
  \centering
  \includegraphics[width=0.38\linewidth]{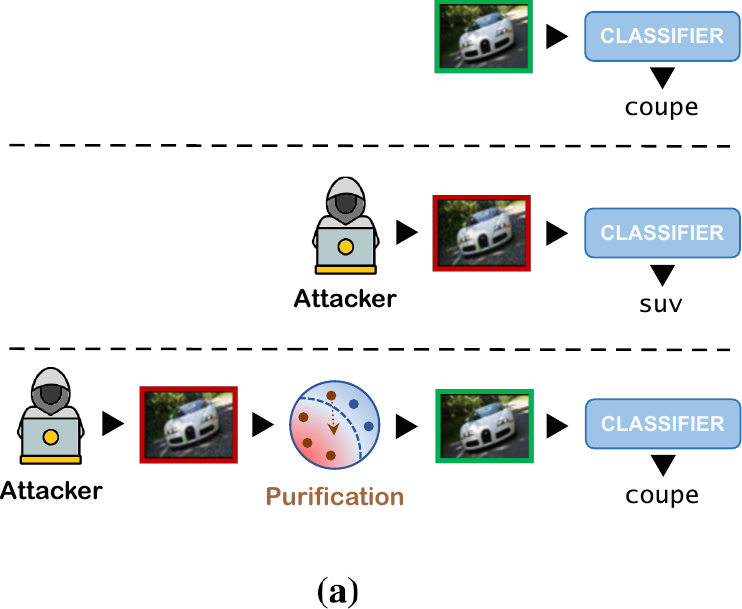}
  \hfill
  \includegraphics[width=0.54\linewidth]{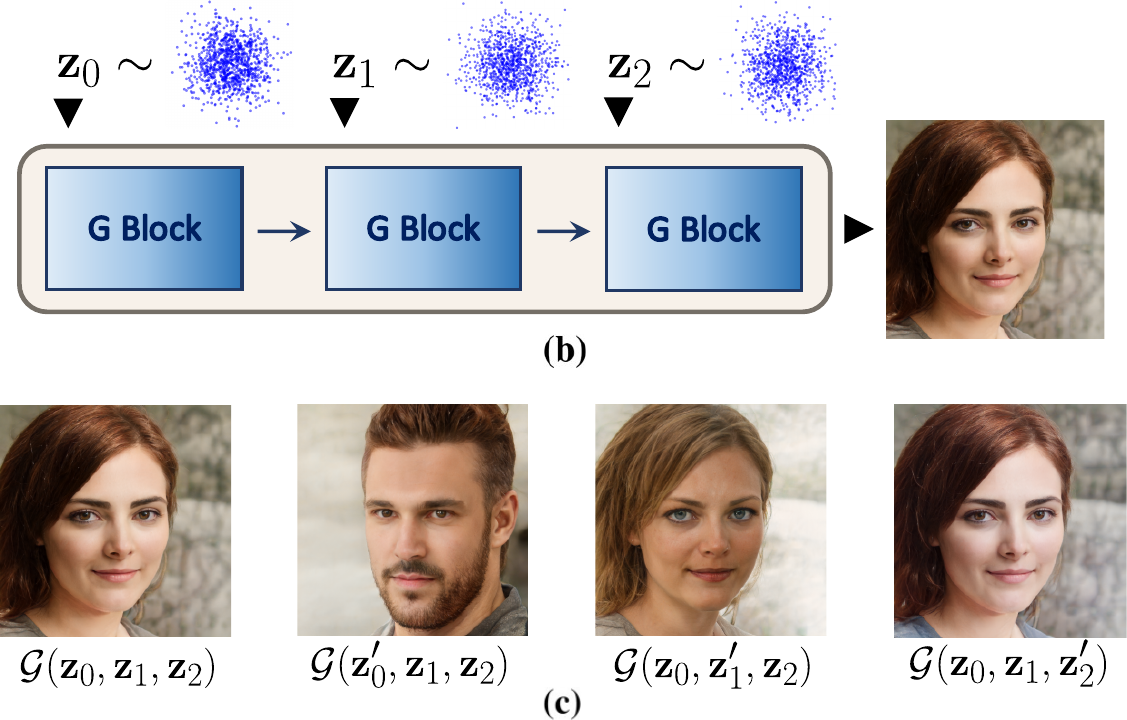}
  \captionof{figure}{\textbf{(a)}: Overview of the adversarial attack and purification mechanism. The attacker subtly perturbs a source image to alter its prediction label (top to center). Purification (bottom) seeks to correct adversarial examples to the right class without affecting clean samples. \textbf{(b)}: A Multiple Latent Variable Generative Model (MLVGM) maps latent variables (or codes, here $\mathbf{z}_0, \mathbf{z}_1, \mathbf{z}_2$) sampled from a known prior distribution, to high-quality images. \textbf{(c)}: Each code impacts the image differently, from global to local features. From left to right: original image and those generated by replacing variables $\mathbf{z}_0$ with $\mathbf{z}_0'$, $\mathbf{z}_1$ with $\mathbf{z}_1'$, and  $\mathbf{z}_2$ with $\mathbf{z}_2'$, respectively.}
  \label{fig:1}
\end{center}%
}]

\begin{abstract}
  \vspace{-10pt}
  Attackers can deliberately perturb classifiers' input with subtle noise, altering final predictions. Among proposed countermeasures, adversarial purification employs generative networks to preprocess input images, filtering out adversarial noise. 
  In this study, we propose specific generators, defined Multiple Latent Variable Generative Models (MLVGMs), for adversarial purification. These models possess multiple latent variables that naturally disentangle coarse from fine features. Taking advantage of these properties, we autoencode images to maintain class-relevant information, while discarding and re-sampling any detail, including adversarial noise. The procedure is completely \emph{training-free}, exploring the generalization abilities of pre-trained MLVGMs on the adversarial purification downstream task.
  Despite the lack of large models, trained on billions of samples, we show that smaller MLVGMs are already competitive with traditional methods, and can be used as foundation models. Official code released at \href{https://github.com/SerezD/gen_adversarial}{https://github.com/SerezD/gen\_adversarial}.
\end{abstract}

\vspace{-10pt}
\section{Introduction}
\label{sec:intro}

It is well known that subtle alterations of input data can significantly impact the predictions of properly trained target models, such as image classifiers. 
This was studied in the pioneering work of Szegedy et al. \cite{adversarial_first} and in follow-up works like \cite{adversarial_second,FSGM_LInf,carlini_wagner}.
An attacker can thus intentionally design an adversarial example with this objective (\Cref{fig:1} (a), top to center), threatening the deployment of deep learning in real-world applications, like smartphone's captured images classification \cite{real_world02} or street sign recognition \cite{real_world01}, to cite a few. 
These vulnerabilities led to an ongoing race between increasingly sophisticated attacks and possible defenses \cite{survey2018,survey2022,survey2023,survey2024}. In fact, researchers proposed different solutions to overcome these shortcomings, like increasing classifier's regularization by training with adversarial images (adversarial training) \cite{adversarial_first,adversarialtraining02,adversarialtraining03,trades}, or applying adversarial purification algorithms \cite{magnet,pixeldefend,defensegan,puvae}. Purification methods (\Cref{fig:1} (a), bottom) act as filters between input images and classifiers, removing adversarial noise. Early approaches (\eg \cite{pixeldefend}) demonstrated that adversarial noise moves the target image to low-confidence regions of the classifier's learned manifold. Therefore, they employed a pre-trained generative model \cite{pixelcnn} as a noise-removal, shifting adversaries back to high-confidence areas. Recently, pre-trained diffusion models \cite{diffusion} have been proposed for the same task \cite{diff_defense01,diff_defense02}, removing noise directly in the pixel space. However, it was shown that specific counter-attacks can prevent their effectiveness \cite{diffattack}.\\
Alternatively, purification can occur by projecting images onto some intermediate latent space \cite{mainfold-vae,nd-vae}. These approaches leverage the regularization properties of VAEs to ensure that close points on the low-dimensional manifold produce similar high-quality images, preserving the coarse semantic content (which determines the category label), while altering local details, including adversarial noise. However, these autoencoder-purification methods are usually specifically trained for the task, significantly increasing the computational overhead.\\
\newline
In this study, we propose a novel autoencoder-purification method, that effectively exploits the power of latent regularizations, but \textit{does not require specific training}. Instead, we define and leverage an emerging class of pre-trained generators, namely, Multiple Latent Variable Generative Models (MLVGMs). Differently from standard latent variable generators like VAEs \cite{vae1,vae2} and GANs \cite{gan}, these models use multiple latent variables (also called \emph{latents} or \emph{codes}) to inject progressive noise during decoding, generating richer and more detailed samples (\Cref{fig:1} (b)). Notable examples include StyleGANs \cite{style-gan,style-gan2,style-gan3,Style-GAN-XL}, GigaGAN \cite{Giga-GAN}, and NVAE \cite{nvae}. As observed in these works, the multiple variables enhance control over the generative process, naturally disentangling global/coarse and local/fine features (see \Cref{fig:1} (c)). So far, these natural disentangling properties have been exploited in various image editing applications, particularly by coupling StyleGAN networks with appropriate encoders \cite{E4E,Transtyle,inversion-1,inversion-2,inversion-3,style-edit}. In this work instead, we take advantage of these properties to perform purification of adversarial images.\\ 
From an MLVGMs perspective, we explore their potential as powerful foundation models, exploiting their abilities in image generation on the zero-shot task of adversarial purification. Foundation models, such as CLIP \cite{clip} or BERT \cite{bert}, are nowadays seen as an easy-to-deploy solution to various downstream tasks, thanks to their generalization capabilities \cite{foundation_models}. Unfortunately, an open-source MLVGM, trained on huge amounts of data and thus serving as a proper foundation model, is not available yet. Interestingly, we find that smaller models, such as StyleGan2 \cite{style-gan2} or NVAE \cite{nvae}, are already powerful enough to be used as purification methods, with no further fine-tuning required. We hope that our findings will set the base for more powerful MLVGMs, trained on billions of samples and deployable on further and diverse downstream tasks.\\
\newline
The motivation for our approach arises from the fact that adversarial images are designed to closely resemble the reference, ``clean'' sample, enforcing an $\textit{L}_{\textit{p}}$ norm bound (usually $\textit{p} \in \{ 0, 2, \hspace{3pt} $or$ \hspace{3pt} \infty \}$). In the pixel space, this imperceptible noise overlaps with class-relevant information, moving the sample to a low-probability region of the targeted classifier's learned manifold \cite{pixeldefend}. Conversely, when encoded in the multiple latent manifolds of MLVGMs, these two types of information likely reside on different levels, since class-relevant information usually has a global impact on the image content, while adversarial information is enforced to change imperceptible, local details. 
Given these observations, we do not limit to remove adversarial noise, as in diffusion-based methods \cite{diff_defense01,diff_defense02}. Instead, we aim to preserve the class-relevant features (essential to determine the final label), while discarding \emph{any} remaining information, including adversarial noise. Since the discarded information does not alter the class label, it can be re-sampled using the generative part of the model, producing clean details.\\
\newline
Our algorithm can be described in three main steps: encoding, sampling, and interpolation. Given an unknown image (adversarial or not), we first \emph{encode} it to obtain $N$ latent variables $\mathbf{z}^{\text{e}}_0, \mathbf{z}^{\text{e}}_1, \dots, \mathbf{z}^{\text{e}}_{N-1}$. These contain relevant class information, which we want to keep, and possible adversarial information, which we need to discard. Second, we \emph{sample} $N$ new latent variables $\mathbf{z}^{\text{s}}_0, \mathbf{z}^{\text{s}}_1, \dots, \mathbf{z}^{\text{s}}_{N-1}$ from the pre-trained generator's prior distribution, as if we would like to synthesize a novel image. Since the MLVGM has learned to represent a clean data distribution during training, these codes do not contain adversarial noise. However, the specific class-relevant information of the codes is unknown. In the third step, we linearly \emph{interpolate} each $\mathbf{z}^{\text{e}}_i$ and $\mathbf{z}^{\text{s}}_i$ according to a certain $0 \le \alpha_i \le 1$.
This is the core of the purification algorithm: we would like to assign low $\alpha_i$ to latent levels representing class-relevant information, thus maintaining the original $\mathbf{z}^{\text{e}}_i$, and high $\alpha_i$ to the latent levels representing irrelevant details, thus practically using the new clean code $\mathbf{z}^{\text{s}}_i$. Finally, we use the generative model to decode the interpolated codes, obtaining the purified image.\\
\newline
With this framework, we leverage the power of VAE-based purification methods, but employ pre-trained MLVGMs, resulting in a training-free procedure and removing the additional overhead. The only adjustable parameters are $\alpha_0, \alpha_1, \dots, \alpha_{N-1}$ which, given a specific problem (target-model and dataset), can be estimated via optimization algorithms, such as Bayesian Optimization (BO) \cite{bayes_opt1,bayes_opt2}. However, since the global, class-relevant information (which we need to maintain) is often contained in the first codes (see \Cref{fig:1} (c)), we show that it is possible to obtain good values of hyperparameters also with reasonable heuristics, by maintaining encoder-information in the first codes while changing the later ones.\\
\newline
We test our framework employing two different MLVGMs (StyleGan2 \cite{style-gan2}, coupled with appropriate encoders \cite{E4E,Transtyle}, and NVAE \cite{nvae}) on three different scenarios: binary classification (male, female) and fine-grained identity classification ($100$ classes) on the Celeb-A dataset \cite{celeba}, and Cars type classification ($4$ classes) on a subset of the Stanford Cars dataset \cite{stanfordcars}. In each scenario, we apply state-of-the-art attacks (DeepFool \cite{deepfool} and Carlini\&Wagner \cite{carlini_wagner}) to compare performances of our model with the base (undefended) classifier, adversarial learning \cite{trades}, and similar generative autoencoding purification methods, namely A-VAE \cite{a-vae} and ND-VAE \cite{nd-vae}. The experimental results show that, despite the used MLVGMs are not originally trained on very large amounts of data, as proper foundation models, they can already compete at the same level of specifically designed techniques, while being completely training-free.\\
\newline
To summarize, our contributions are: 
1) We propose a novel autoencoding-based purification framework, which employs an off-the-shelf pre-trained architecture, namely, Multiple Latent Variable Generative Models (MLVGMs), requiring at most the estimation of a handful of hyperparameters;
2) We show the potential of MLVGMs as strong foundation models for adversarial purification:
thanks to the regularization and disentangling properties of the multiple latent spaces, these models are proved to be effective for downstream tasks without the need to be specifically trained.
3) Our experiments encompass different image domains and MLVGMs, setting them as a valid alternative to other purification methods, regardless of the specific training procedure (VAE or GAN) and despite not being pre-trained on billions of samples, as proper foundation models.

\section{Background and Related Works}

In this section, we provide the minimal background on adversarial attacks, before delving into the literature on adversarial defenses and MLVGMs.

\label{sec:related_works}

    \paragraph{Adversarial attacks.}

        As noted in \cite{adversarial_first}, subtle changes to a classifier's input can lead to incorrect predictions. These often imperceptible perturbations do not affect the input's semantic content, revealing vulnerabilities in the model's robustness and security. Formally, these adversarial perturbations are defined as: 
        \begin{equation}
        \label{eq:attack}
            \argmin_\delta \| \delta \| \text{ s.t. } \mathbb{I}(f(\mathbf{x} + \delta) \ne \mathbf{y}),
        \end{equation}
        where $f(\mathbf{x} + \delta)$ is the prediction of the model $f(\cdot)$ for the input $\mathbf{x}$ with perturbation $\delta$, $\mathbf{y}$ is the ground truth label, and $\mathbb{I}(\cdot)$ is the indicator function. Opposed to the untargeted case of \Cref{eq:attack}, an attack is targeted when $f(\mathbf{x} + \delta)$ is forced to output a specific incorrect label $\mathbf{y}' \ne \mathbf{y}$.
        To measure the perturbation and evaluate attack effectiveness, $\textit{L}_{\textit{p}}$ norm is used, where $\textit{p} \in \{ 0, 2, \infty\}$. An attack is considered successful within the bound $\epsilon$ if the minimal perturbation found satisfies $\|\delta\|_{\textit{p}} \le \epsilon$.\\ 
        \newline
        Based on the attacker's knowledge of the target model, adversarial attacks can be categorized into \textit{blackbox} and \textit{whitebox}. The former class \cite{boundary,hsja}  relies only on the model's final predictions, while the latter \cite{FSGM_LInf,PGD,BrendelBethge} assumes full access to the model. For a complete overview of the numerous existing attacks, we refer to recent surveys like \cite{survey2018,survey2022,survey2023,survey2024}.
        In this work we test against \textit{whitebox} and untargeted attacks, representing the most challenging setup for a defense mechanism. We select two different attacks, DeepFool \cite{deepfool} and Carlini-Wagner (C\&W) \cite{carlini_wagner}, as representative state-of-the-art methods employing different techniques to find the best perturbation. DeepFool iteratively perturbs the input image to find the minimal change required to alter the classification result. At each step, it linearly approximates the decision boundary and projects the image toward it until misclassification occurs. Conversely, C\&W formulates the attack as an optimization problem, utilizing gradient information to minimize the perturbation while maximizing classifier's loss.

    \paragraph{Defending against adversarial attacks.}

       The first proposed approach to reduce the effectiveness of attacks is known as adversarial training \cite{adversarial_first,FSGM_LInf,adversarialtraining02,adversarialtraining03}, which involves generating adversarial examples during the training phase to enhance the classifier's robustness. Although scalable to large datasets \cite{adversarialtraining04}, this method has drawbacks, including a significant computational cost and the risk of overfitting to specific attacks. Noteworthy is also the area of certified robustness \cite{certified01,certified02,certified03,certified04,certified05} providing theoretical guarantees that any perturbation within a certain $\textit{L}_{\textit{p}}$ norm ball is classified correctly. While promising, these methods often give guarantees only to a specific $\textit{p}$ norm, or excessively reduce the classifier's accuracy on clean data.\\
       \newline
       Our work belongs instead to the well established concept of adversarial purification, which involves removing adversarial noise from input images.  Seminal works such as \cite{magnet,pixeldefend} postulate that adversarial samples reside in low-probability regions of target classifier's manifold, reducing confidence. Thus, the objective of purification is to move adversarial samples back towards high-probability regions, removing the applied noise. This approach acts as a preprocessing filter of input images, maintaining the classifier unaltered and allowing the combination with other techniques. Subsequent studies leveraged the regularization properties of latent variable generative models, including GANs \cite{gan} and VAEs \cite{vae1,vae2}. Specifically, \cite{defensegan} proposed identifying the latent space vector $\mathbf{z}$ of a pre-trained GAN $g_{\pmb{\theta}}$ that minimizes the distance between the real input $\mathbf{x}$ and the generated $g_{\pmb{\theta}}(\mathbf{z})$, which serves as the purified image. Other methods like \cite{puvae,defensevae,nd-vae} train a VAE to reconstruct clean images from adversarial examples. Further VAE-based approaches like \cite{a-vae,mem-vae} hypothesize that adversarial attacks primarily manipulate local information. Therefore, they aim at preserving coarse features while purifying the details. Our method follows this principle, but we leverage the powerful representations of \emph{pre-trained} models, arguing that the training of specific autoencoders is not necessary.
       In general, the use of pre-trained generators has been proposed also in the context of diffusion-based purification \cite{diff_defense01,diff_defense02}. However, these approaches act directly in the pixel-space, and can be countered by specific attacks \cite{diffattack}. Instead, we leverage the regularization and disentanglement properties of MLVGMs latent spaces to discard and re-sample any information that is not relevant for the final class label, increasing the attacker's challenge.\\
       \newline
       Similarly to other pipelines, our method includes random sampling operations. As seen in \cite{obfuscated_gradients}, randomized defenses can give a false sense of robustness, by masking true gradients to the attacker. Gradient obfuscation can be easily circumvented by averaging the gradients over multiple input transformations, taking the Expectation over Transformation (EoT)  \cite{eot}. In our experimental protocol, we consider this aspect and appropriately use EoT to avoid gradient masking.
       
    \paragraph{Multiple Latent Variable Generative Models.}

        Current literature encompasses a large collection of generative models employing multiple random variables, especially those based on GANs \cite{lap-gan,biggan,Giga-GAN} and VAEs \cite{ladder-vae,deep-ladder-vae,vlae,disentangled-vlae,nvae}. Notably, these have been previously leveraged in image editing tasks \cite{style-edit,E4E,Transtyle}, particularly employing the StyleGan family of networks \cite{style-gan,style-gan2,style-gan3,Style-GAN-XL}. Image editing tasks are feasible only when the StyleGan is coupled with an appropriate encoder network, allowing for the extraction of latent codes from real images \cite{inversion-1,inversion-2,inversion-3}, a task known as GAN inversion. In this work, we also couple StyleGan with appropriate encoder networks, but we use the disentanglement properties of coarse from finer information for the task of adversarial purification. By doing so, we show that MLVGMs have the potential to be used as foundation models, despite generators trained on billions of samples, like GigaGAN \cite{Giga-GAN} are not fully available yet. Interestingly, we observe that the field of MLVGMs is expanding, with promising approaches based on Normalizing-Flow \cite{rg-flow}, and SODA \cite{soda}, a diffusion model coupled with a multi-latent encoder.

\section{Methodology}
\label{sec:methodology}
%
\begin{figure*}
  \centering
  \includegraphics[width=0.9\linewidth]{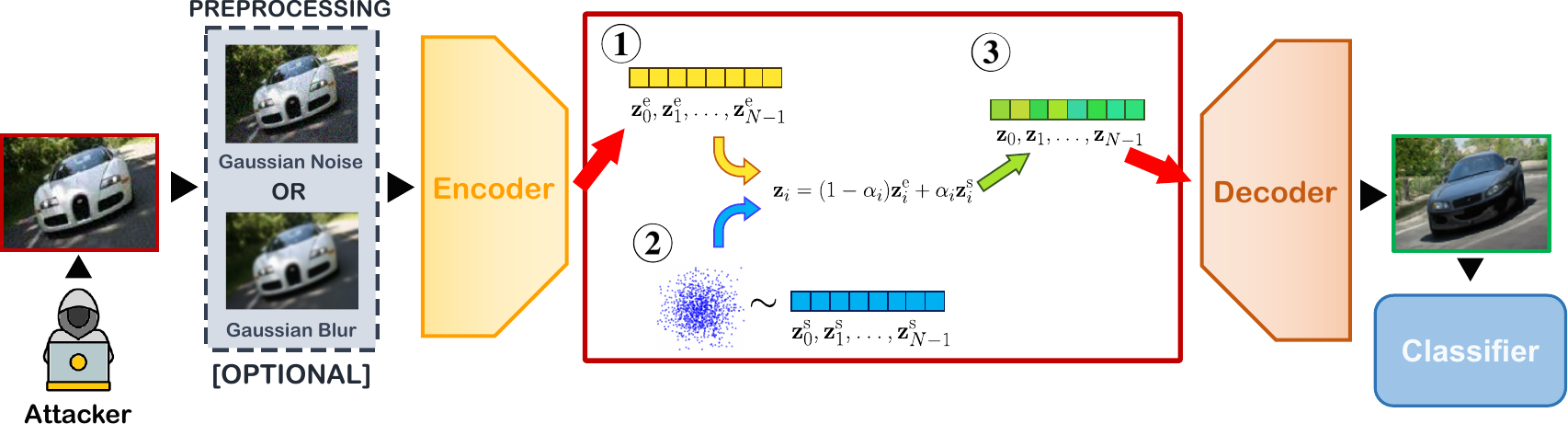}
  \caption{The overall architecture of our framework, depicting the \emph{optional} preprocessing phase (adding gaussian noise or blurring the input image) and autoencoding with pre-trained MLVGMs, which \textit{do not require further training}. Inside the central rectangle, we show the latent purification process that is the core of our method, consisting of three main steps: \circled{1} encoding, \circled{2} sampling and \circled{3} interpolation.}
  \label{fig:2}
  \vspace{-15pt}
\end{figure*}
\paragraph{Purification Framework with MLVGMs.} 

The proposed framework's overview is shown in \Cref{fig:2}, while we provide the pseudocode in \Cref{supp:pseudocode}. Initially, the input image $\mathbf{x}$ undergoes an \emph{optional} preprocessing stage, consisting of noise enhancement (adding Gaussian noise) or suppression (applying Gaussian blur). Similarly to previous methods like \cite{nd-vae,a-vae}, we experimentally observe that this step can sometimes increase the challenge for the attacker. Even tough our method works even with no preprocessing on the input images, these optional operations demonstrate that it can be successfully coupled with other pipelines to increase overall performance, while  maintaining its stability.\\
\newline
After the initial preprocessing, we leverage the pre-trained generative autoencoder for the core purification procedure (big rectangle in \Cref{fig:2}). First, we encode the input image to obtain multiple latent variables $\mathbf{z}^{\text{e}}_0, \mathbf{z}^{\text{e}}_1, \dots, \mathbf{z}^{\text{e}}_{N-1}$ (\Cref{fig:2}, \circled{1}). These codes represent different levels of information, from coarse to fine, capturing both class-relevant features (to be maintained) and irrelevant details/adversarial noise (to be suppressed). Next, we sample $N$ new latent codes $\mathbf{z}^{\text{s}}_0, \mathbf{z}^{\text{s}}_1, \dots, \mathbf{z}^{\text{s}}_{N-1}$ from each latent prior distribution, utilizing the model's generative properties (\Cref{fig:2}, \circled{2}). These latent variables form a novel image with an unknown final classification label. However, since the generative model is trained to represent the real data distribution accurately, the sampled codes do not contain adversarial information.\\
\newline
Given the two types of latent variables ($\mathbf{z}^{\text{e}}_i$ and $\mathbf{z}^{\text{s}}_i$), we aim to obtain a new set of codes, retaining the class-relevant features of $\mathbf{z}^{\text{e}}_i$ and the clean details of $\mathbf{z}^{\text{s}}_i$. Therefore, we linearly interpolate each $\mathbf{z}^{\text{e}}_i$ and $\mathbf{z}^{\text{s}}_i$ (\Cref{fig:2}, \circled{3}):
\begin{equation}
\mathbf{z}_i = (1 - \alpha_i) \mathbf{z}^{\text{e}}_i + \alpha_i \mathbf{z}^{\text{s}}_i,
\end{equation}
where $0 \le \alpha_i \le 1$, with $\alpha_i = 0$ using only the encoding information and $\alpha_i = 1$ using only the new information. The outcome of this interpolation process is $N$ purified codes $\mathbf{z}_0, \mathbf{z}_1, \dots, \mathbf{z}_{N-1}$, which, once decoded, produce a purified version of the input image, becoming the classifier's input (rightmost part of \Cref{fig:2}). The crucial aspect of this process, discussed next, resides in the selection of the optimal $\alpha_0, \alpha_1, \dots, \alpha_{N-1}$ to maintain the input's class-relevant information while effectively removing anything else.
As seen in the purified image in \Cref{fig:2}, we do more than just remove adversarial noise. The goal is to find the parameters that alter \emph{any} features irrelevant to class labels, such as the car's color or background. By maximizing the discarded and resampled information, we restrict the attacker's degrees of freedom and effectiveness. For instance, if only $\mathbf{z}^e_0$ is retained after purification, to affect the final classifier the attacker is \emph{forced} to alter parts of the input that are encoded at that level. However, since in MLVGMs the first codes typically influence global parts of the image, this constraint struggles with the enforced $\textit{L}_{\textit{p}}$ bound on adversarial noise, which affects only imperceptible details. This double constraint (one in the latent space, one in the pixel space), poses a great challenge for the attacker, limiting the overall effectiveness of the perturbation.

\paragraph{Selection of $\alpha$ hyperparameters.}

We now discuss possible strategies to find the optimal choice of the $\alpha$ parameters, which should maintain class-relevant information from $\mathbf{z}^{\text{e}}$ codes and introduce novel details thanks to  $\mathbf{z}^{\text{s}}$. Specifically, we propose using Bayesian Optimization \cite{bayes_opt1,bayes_opt2}, an effective algorithm for finding the best hyperparameters in machine learning when the search space is small and continuous. For $\alpha$ selection, each combination has $N$ dimensions, corresponding to the number of latent levels (up to 24 in our experiments), with ${\alpha \in \mathbb{R}^N : 0 \le \alpha_i \le 1}$. Bayesian Optimization is also suitable when evaluating the \emph{objective function} is complex, making simpler algorithms like grid search impractical. In our setup, the \emph{objective function} evaluates purification performance for a given set of hyperparameters. Specifically, we define a ``base model'' as the framework described in \Cref{fig:2}, without preprocessing, and with $\alpha_i = 0, \forall i$, corresponding to simple autoencoding $+$ classification. Then, we attack this model by applying FGSM \cite{FSGM_LInf} to every image in the source dataset, obtaining its adversarial version. We define the \emph{objective function} of Bayesian Optimization as the accuracy computation of the base model on the adversarial dataset. At each step, we change the set of hyperparameters to match the one suggested by the optimization algorithm.\\
\newline
The use of optimization algorithms, however, presents some drawbacks: the requirement of specific adversarial images to compute the \emph{objective function} and the overall computational cost. In fact, evaluating the \emph{objective function} may still be infeasible in resource-constrained scenarios with very limited resources. In other terms, it would be desirable to have a complete optimization-free purification, using pre-trained MLVGMs without hyperparameters tuning. In the following, we propose such an alternative, using the properties of MLVGMs to define what characteristics a good combination should have.\\
\newline
As discussed in \Cref{sec:intro}, MLVGMs encode information at various granularities, from global to local details (\Cref{fig:1} (\textbf{c})), a behavior observed in studies like \cite{biggan,nvae,style-gan}. In most classification problems, global aspects are more relevant for label definition than pixel-level details. For example, in identity classification, general face traits are more crucial than localized features, which can change without affecting the label. Based on this insight, we hypothesize that a good set of $\alpha$ parameters should retain encoding information from initial latent levels while replacing the later ones with the sampled codes. Thus, when optimization is infeasible, we propose selecting monotonic sets of values, where $\alpha_i > \alpha_j$ for all $i > j$. Specifically, we experiment with two methods:

{\small
\vspace{-10pt}
\begin{align}
\label{eq:linear}
\text{\textbf{linear}} & \quad \alpha_i = \frac{(i + 1)}{N} &  \forall i \in \{0, 1, \dots, N-1\}; \\
\label{eq:cosine}
\text{\textbf{cosine}} & \quad \alpha_i = \frac{1 - \cos{\frac{\pi (i + 1)}{N}}}{2}  & \forall i \in \{0, 1, \dots, N-1\};
\end{align}
}

\noindent finding them to be stable across various scenarios, and competitive with the combination found after optimization.

\section{Experiments}

\label{sec:experiments}

\paragraph{Pre-trained MLVGMs, datasets and classifiers.}

Unfortunately, no proper foundation MLVGM (trained on billions of images) is available yet. GigaGAN \cite{Giga-GAN} shows impressive results, but no pre-trained model has been released. Therefore,   
we experiment with three types of smaller MLVGMs: a StyleGan2 \cite{style-gan2} model pre-trained on the Celeb-A HQ dataset \cite{celeba}, coupled with the E4E Encoder proposed by \cite{E4E}; an NVAE model \cite{nvae}, also pre-trained on Celeb-A; and a StyleGan2 model pre-trained on LSUN Cars \cite{lsun}, coupled with the Style Transformer Encoder (STE) proposed by \cite{Transtyle}. 
To further enhance the broadness of our experiments, each model is tested for its purification capability on a dedicated classification problem, considering the inherent drawbacks of each.\\
\newline
The E4E-StyleGan2 model suffers from imperfect reconstructions, due to the post hoc training of the encoder. Therefore, we choose a binary classification task (male/female), ensuring that the relevant global aspects (gender) are maintained after reconstruction. Specifically, we employ the gender classification dataset introduced in \cite{celeba_gender}, containing $24$k training and $6$k validation images, at a resolution of $256 \times 256$. Next, we train an NVAE to demonstrate our framework's compatibility also with VAE-based architectures and to tackle a more challenging classification task, possible thanks to precise reconstructions. We used a subset of the Celeba-Identities \cite{celeba} dataset, comprising $2600$ training and $600$ validation images, on $100$ identity classes and a $64\times 64$ resolution. Lastly, for the STE-StyleGAN2 model we select cars classification, to show applicability across different image domains. For the same reasons as E4E-StyleGan2, also STE-StyleGAN2 suffers from imperfect reconstructions. Therefore, we use a subset of the Stanford Cars dataset \cite{stanfordcars}, grouping the provided label names into four classes: Coupe, Hatchback, SUV and Minivan. The dataset contains $800$ $128 \times 128$-resolution images per class, keeping $100$ of them for validation.\\
\newline
As classifiers, we trained a Resnet-50 \cite{resnet50} for $50$ epochs, a Vgg-11 \cite{vgg11} for $200$ epochs and a Resnext-50 \cite{resnext32} model for $150$ epochs, respectively. All use SGD optimizer and a learning rate of $1e-3$. We provide more details and hyperparameters in \Cref{supp:exp_details}.

\paragraph{Threat model and baselines. }
As anticipated in \Cref{sec:related_works}, we evaluate each purification model under the challenging scenario of \textit{whitebox} untargeted attacks, constrained by an $\text{L}_2$ norm. This means that the attacker has complete access to the entire model, including the purification framework, and the attack is deemed successful if the target classifier predicts \emph{any} incorrect class. Specifically, we test our model against two prominent adversarial attacks: DeepFool \cite{deepfool} and Carlini \& Wagner (C\&W) \cite{carlini_wagner}\footnote{Further experiments on Autoattack \cite{autoattack} are provided in \Cref{supp:autoattack}.}. To obviate the problem of gradient obfuscation, we couple each attack with Expectation over Transformation (EoT) \cite{eot}. In all cases, we average gradients over $32$ forward passes, or the maximum amount allowed by the available resources.\\
\newline
Given an attack, we seek for each sample the minimal perturbation needed to cause a misclassification, obtaining an adversarial set of images $\{\hat{\mathbf{x}}_i^{\delta_i} \}_{i=1}^N$ where $N$ is the size of the test set, $\delta_i$ is the minimum perturbation found for sample $\mathbf{x}_i$ and $\hat{\mathbf{x}}_i^{\delta_i} = \mathbf{x}_i + \delta_i$. Then, we compute the average success rate (SR) at a certain $\textit{L}_2$ bound as:
\begin{equation}
    \text{SR}_{\textit{L}_2 = \epsilon} = \frac{1}{N} \sum_{i=1}^N  \mathbb{I}((\hat{y}_i \ne y_i) \land (\delta_i \le \epsilon)),
\end{equation} 
where $\hat{y}_i$ is the predicted label of the target model given the sample $\hat{x}_i^{\delta_i}$, $y$ is the ground truth label, $\epsilon$ the considered $\textit{L}_2$ bound and $\mathbb{I}(\cdot)$ the indicator function. In each experiment, we report the success rate obtained on $N = 100$ images uniformly sampled from the available classes. Further details and a comprehensive list of hyperparameters used are included in \Cref{supp:exp_details}.\\
\newline
For each scenario, we first measure the success rate of the different hyperparameter choices (\textbf{learned}, \textbf{linear}, and \textbf{cosine}). Then, we ablate on the effects of adding a preprocessing operation (see \Cref{supp:ablations}). Lastly, we benchmark the best-performing configuration (hyperparameters and preprocessing) against: the classification model alone (no defense), the classification model regularized with adversarial training, using TRADES \cite{trades}, and similar adversarial-purification methods, A-VAE \cite{a-vae} and ND-VAE \cite{nd-vae}. \Cref{tab:hp_costs} shows the cost of each method, in terms of the number of parameters to optimize. For TRADES, we fine-tuned the classifier for $50$ epochs, while for ND-VAE and A-VAE we trained the additional Generative-Autoencoder from scratch, on each dataset (further details in \Cref{supp:exp_details}). Conversely, our method requires optimizing just a few parameters in the \textbf{learned} case, and none if \textbf{linear} or \textbf{cosine} configurations are used.
\begin{table}
  \centering
  {\small{
  \begin{tabular}{@{}lc@{}}
    \toprule
    Method & Optimized Params. \\
    \midrule
    Trades \cite{trades} & $10^7$ \\
    A-VAE \cite{a-vae} & $10^7$\\
    ND-VAE \cite{nd-vae} & $10^6$ \\
    \hline
    ours w/ BO & $10^1$ \\
    ours w/o BO & $0$ \\
    \bottomrule
  \end{tabular}
  }}
  \caption{Average (on each task) order of magnitude of parameters that need to be optimized for each compared method.}
  \label{tab:hp_costs}
  \vspace{-15pt}
\end{table}
\paragraph{Effects of $\alpha$ configurations.}
\begin{figure*}
  \centering
    \includegraphics[width=1.0\linewidth]{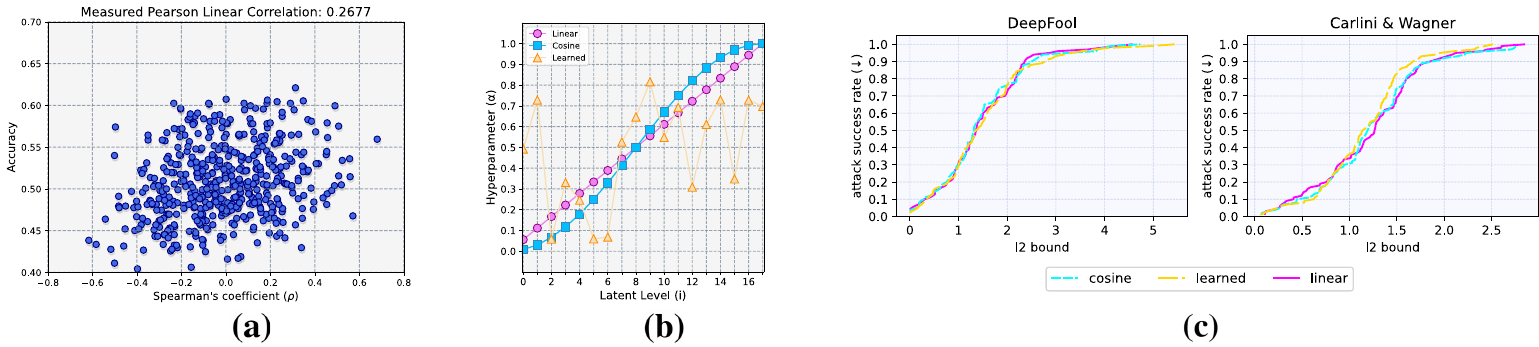}
    \vspace{-15pt}
    \caption{Analyisis of $\alpha$ combinations on the Celeb-A HQ Gender task. \textbf{(a)} Spearman's index ($\rho$) vs accuracy for $512$ random combinations, obtaining a Pearson's linear correlation value of $0.267$. \textbf{(b)} Comparison of the $18$ final $\alpha$ values for the \textbf{linear}, \textbf{cosine} and \textbf{learned} combinations. \textbf{(c)} Attack success rates (the lower the better) for increasing $\textit{L}_2$ bounds on each tested attack and combination.}
    \label{fig:gender_combinations}
\end{figure*}
\begin{figure*}
  \centering
    \includegraphics[width=1.0\linewidth]{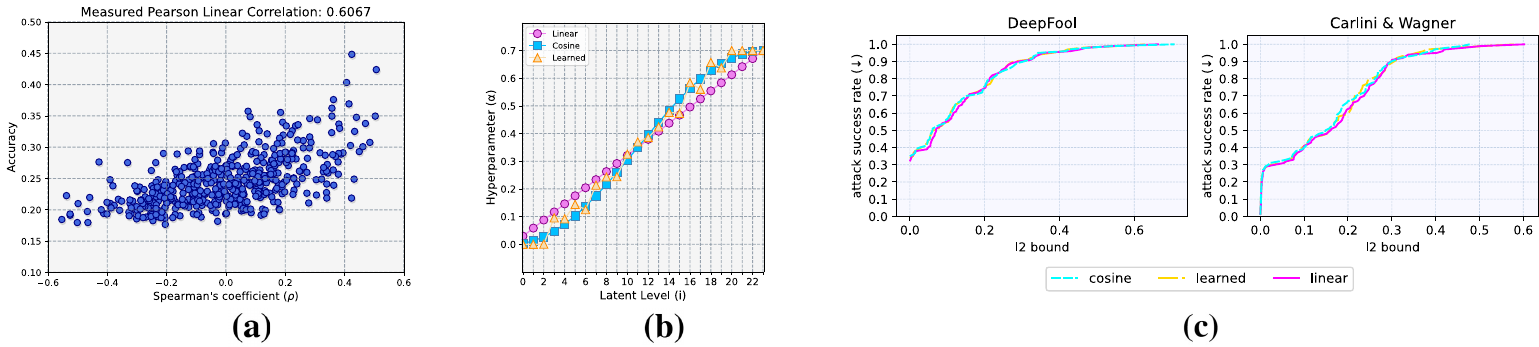}
    \vspace{-15pt}
    \caption{Analyisis of $\alpha$ combinations on the Celeb-A 64 Identities task. \textbf{(a)} Spearman's index ($\rho$) vs accuracy for $512$ random combinations, obtaining a Pearson's linear correlation value of $0.607$. \textbf{(b)} Comparison of the $24$ final $\alpha$ values for the \textbf{linear}, \textbf{cosine} and \textbf{learned} combinations. \textbf{(c)} Attack success rates (the lower the better) for increasing $\textit{L}_2$ bounds on each tested attack and combination.}
    \label{fig:ids_combinations}
    \vspace{-12pt}
\end{figure*}
\begin{figure*}
  \centering
    \includegraphics[width=1.0\linewidth]{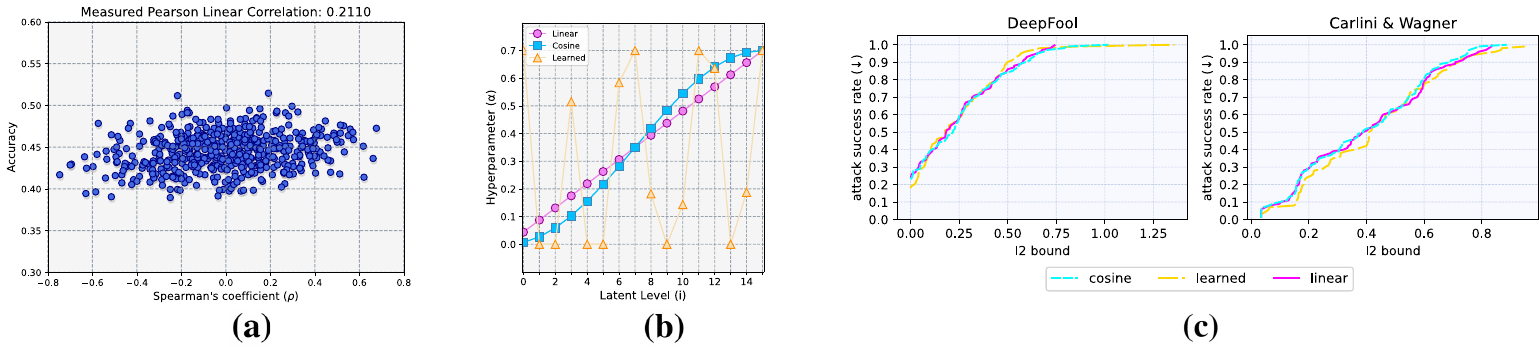}
    \vspace{-15pt}
    \caption{Analyisis of $\alpha$ combinations on the Stanford Cars 128 task. \textbf{(a)} Spearman's index ($\rho$) vs accuracy for $512$ random combinations, obtaining a Pearson's linear correlation value of $0.211$. \textbf{(b)} Comparison of the $16$ final $\alpha$ values for the \textbf{linear}, \textbf{cosine} and \textbf{learned} combinations. \textbf{(c)} Attack success rates (the lower the better) for increasing $\textit{L}_2$ bounds on each tested attack and combination.}
    \label{fig:cars_combinations}
    \vspace{-12pt}
\end{figure*}
To offer a broader perspective on the effectiveness of choosing a monotonic set of hyperparameters, we propose to measure it statistically. Specifically, we use the ``base model'' detailed in \Cref{sec:methodology} to measure the accuracy of $512$ random combinations of $\alpha$ values, sampled uniformly. For each combination, we also calculate the Spearman's rank correlation index \cite{spearman}, to gauge its monotonicity degree:
\begin{equation}
    \rho = 1 - \frac{\sum_{i=0}^{N-1} (i - R_{\alpha_i})^2}{N(N^2 - 1)};
\end{equation}
where $N$ is the number of latent levels and $R_{\alpha_i}$ is the rank of the parameter $\alpha_i$, ranging from $0$ to $N-1$. We plot the two variables (Spearman's rank vs Accuracy) for each task in \Cref{fig:gender_combinations,fig:ids_combinations,fig:cars_combinations} (a), and measure their Pearson linear correlation coefficient \cite{pearson}. We obtain values of $0.267, 0.607, 0.211$ respectively, where a value close to one indicates that a monotonic set of hyperparameters is particularly effective. In all cases, linear correlation exhibits  a positive value, meaning that a relation between monotonicity and accuracy exists. This is particularly strong in the fine-grained problem of ids classification. We hypothesise that in this case relevant class-features are spread across various latents, in a coarse to fine manner. Thus, a gradually increasing set of $\alpha$ values is particularly effective. On the other hand, in coarse-grained classification, relevant features are mainly concentrated in a few, initial latents, implying that a monotonic set of values performs well, but is not essential. Therefore, learning the best combination may give better results in such cases.\\
\newline
To learn the best combination with Bayesian Optimization, we use the BoTorch library \cite{botorch} and fit a Gaussian Process model for $95$ steps, using Expected Improvement as the acquisition function. Five additional combinations are given for initialization: \textbf{linear} (\cref{eq:linear}), \textbf{cosine} (\cref{eq:cosine}), uniform ($\alpha_i = 0.5, \forall i \in \{0, 1, \dots, N-1\}$), $1 - \text{linear}$, $1 - \text{cosine}$. In \Cref{fig:gender_combinations,fig:ids_combinations,fig:cars_combinations} (b) we visually compare the \textbf{learned} combination with the fixed ones. In ids and cars tasks, we force the maximum $\alpha$ value to $0.7$, since higher values caused an high degradation of clean accuracy on \textbf{linear} and \textbf{cosine} cases. Aligning with previous observations, the best results for ids-classification are obtained with a monotonic set of values, while in other cases BO highly penalizes some latent levels, maintaining information unaltered in other ones. We qualitatively analyze these \textbf{learned} combinations in \cref{supp:preliminary_studies}. \Cref{fig:gender_combinations,fig:ids_combinations,fig:cars_combinations} (c) shows the purification abilities on the DeepFool and C\&W attacks. The combination learned on FGSM for gender classification works well on DeepFool, but no on C\&W. On ids-classification, all combinations unsurprisingly perform similarly. For cars-classification, the \textbf{learned} combination allows an extra performance boost on both attacks.

\paragraph{Comparison with other methods.}
\begin{figure}
  \centering
  \includegraphics[width=1.0\linewidth]{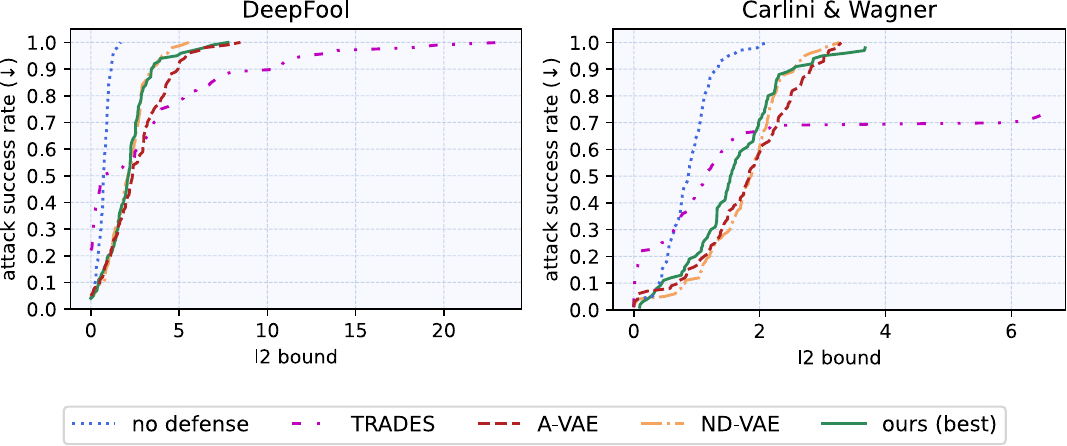}
  \caption{Attack success rates (lower is better) for increasing $\textit{L}_2$ bounds. Comparison of different defenses on Celeb-A HQ Gender.}
  \label{fig:results_gender}
  \vspace{-5pt}
\end{figure}
\begin{figure}
  \centering
  \includegraphics[width=1.0\linewidth]{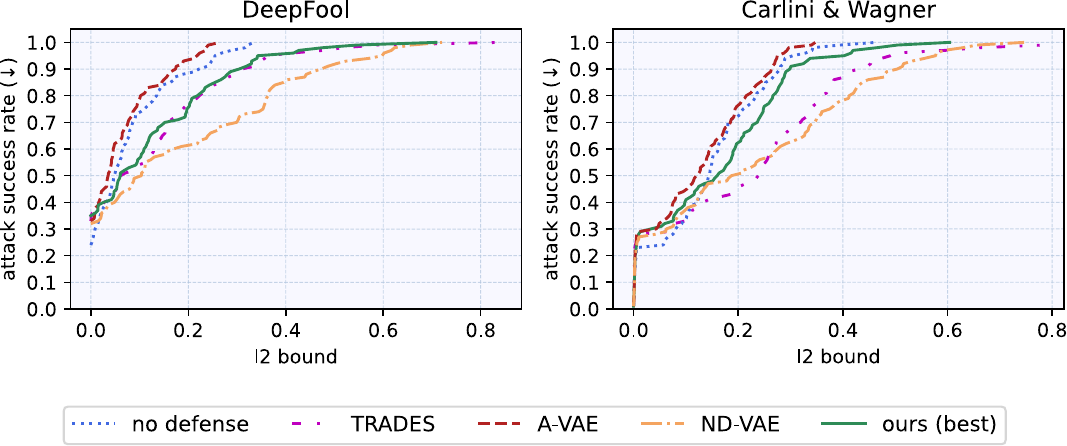}
  \caption{Attack success rates (lower is better) for increasing $\textit{L}_2$ bounds. Comparison of different defenses on Celeb-A 64 Identities.}
  \label{fig:results_ids}
  \vspace{-5pt}
\end{figure}
\begin{figure}
  \centering
  \includegraphics[width=1.0\linewidth]{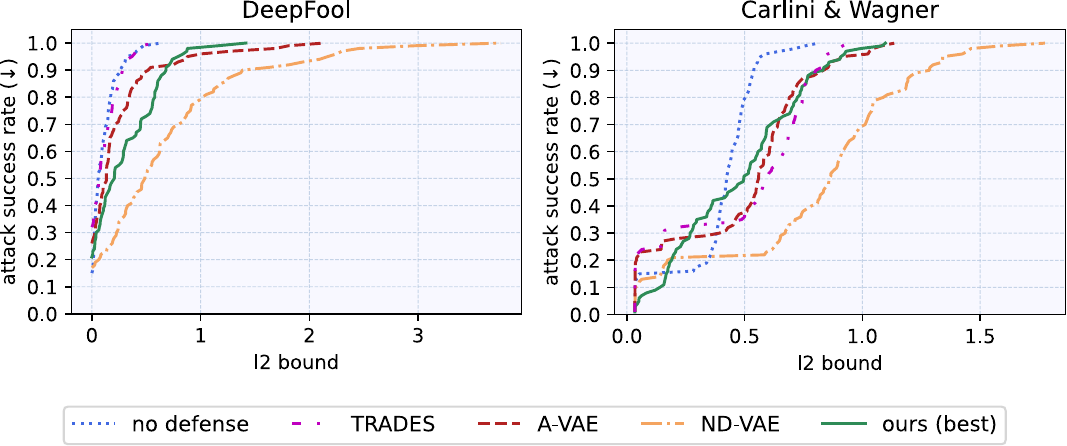}
  \caption{Attack success rates (lower is better) for increasing $\textit{L}_2$ bounds. Comparison of different defenses on Stanford Cars 128.}
  \label{fig:results_cars}
  \vspace{-15pt}
\end{figure}
\Cref{fig:results_gender,fig:results_ids,fig:results_cars}
show the comparison between the best configuration of our framework and similar defense methods. In \Cref{supp:ablations} we ablate on pre-processing operations, finding that gaussian blur is beneficial on gender classification, while gaussian noise performs best on the other tasks. We refer to \Cref{supp:qualitative_examples} for some qualitative purification examples. Looking at the figures, TRADES shows very strong results on gender classification, outlining the power of Adversarial Training in 2-classes problems; while all purification methods (including ours) perform similarly. In the more challenging scenarios of ids and cars classification, conversely, ND-VAE generally performs best, with our method always showing competitive results. More broadly, MLVGMs prove to be good adversarial purifiers, despite not being specifically trained for the task and the relatively small size of the used models. We hypothesise that this is due to the coarse-to-fine disentenglement of their features, which offers promising reasearch directions for the development of more powerful MLVGMs, acting as proper foundation models. 

\section{Discussion and Conclusions}
\label{sec:conclusion}
In this paper we proposed a novel adversarial purification method, using Multiple Latent Variable Generative Models (MLVGMs) as foundation models. Our approach takes advantage of their latent disentanglement properties, leveraging them on the adversarial purification downstream task in a training-free manner. In the worst case, only a few hyperparameters need to be optimized via Bayesian Optimization. However, thanks to the global-to-local features of MLVGMs, good values can be defined a priori.\\
While promising, the use of MLVGMs as foundation models still poses some challenges. Specifically, the lack of strong open-source models, trained on billions of samples, is a limiting factor in obtaining stronger results. In our study, we employed StyleGan2 \cite{style-gan2} and NVAE \cite{nvae} networks, which are smaller models that present some intrinsic limitations. Despite this, the proposed framework already shows competitive results, when compared to similar defense models that require specific training. Therefore, our study highlights the significant potential of MLVGMs as strong foundation models, encouraging research to release more powerful generators.


\clearpage
\setcounter{page}{1}
\maketitlesupplementary
\appendix

In completion of the main paper, we provide additional information regarding experimental content and qualitative results. This Supplementay Material is organized as follows: \Cref{supp:pseudocode} contains the pseudocode of the purification procedure; \Cref{supp:preliminary_studies} a qualitative analysis of images generated following the \textbf{learned} set of $\alpha$ hyperparameters; \Cref{supp:exp_details} experimental details that can help in reproducing the results; \Cref{supp:ablations} the ablation studies and details on preprocessing operations; \Cref{supp:qualitative_examples} shows some qualitative purification examples of our method, in comparison to A-VAE \cite{a-vae} and ND-VAE \cite{nd-vae}; and \Cref{supp:autoattack} contains additional defense results on the Autoattack method \cite{autoattack}.

\section{Pseudocode}
\label{supp:pseudocode}

In \Cref{alg:purification} we show the pseudocode for the purification procedure, as explained in \Cref{sec:methodology} of the main paper. 

\begin{algorithm}
\caption{Purification Procedure}
\label{alg:purification}
\begin{algorithmic}[1]
    \Require Input Image $\mathbf{x}$; Preprocessing Function $\mathcal{P}$; Encoder and MLVGM $\mathcal{E}, \mathcal{G}$; Generator's Prior $\mathcal{N}$; Hyperparameters $\{\alpha_0, \alpha_1, \dots, \alpha_{N-1}\}$.
    \State $\mathbf{x}_{\text{p}} = \mathcal{P}(\mathbf{x})$ \Comment{optional preprocessing}
    \State $\mathbf{z}^{\text{e}}_0, \mathbf{z}^{\text{e}}_1, \dots, \mathbf{z}^{\text{e}}_{N-1} = \mathcal{E}(\mathbf{x}_{\text{p}})$ \Comment{encoding}
    \For{$i = 0$ to $N-1$}
        \State $\mathbf{z}^{\text{s}}_i \sim \mathcal{N}$ \Comment{sampling}
        \State $\mathbf{z}_i = (1 - \alpha_i) \mathbf{z}^{\text{e}}_i + \alpha_i \mathbf{z}^{\text{s}}_i$ \Comment{interpolation}
    \EndFor
    \State $\hat{\mathbf{x}} = \mathcal{G}(\mathbf{z}_0, \mathbf{z}_1, \dots, \mathbf{z}_{N-1})$ \Comment{decoding}
    \State \Return Purified Image $\hat{\mathbf{x}}$
\end{algorithmic}
\end{algorithm}

\section{Qualitative Analysis of Bayesian Optimization}
\label{supp:preliminary_studies}

\begin{figure*}
  \centering
  \includegraphics[width=1.0\linewidth]{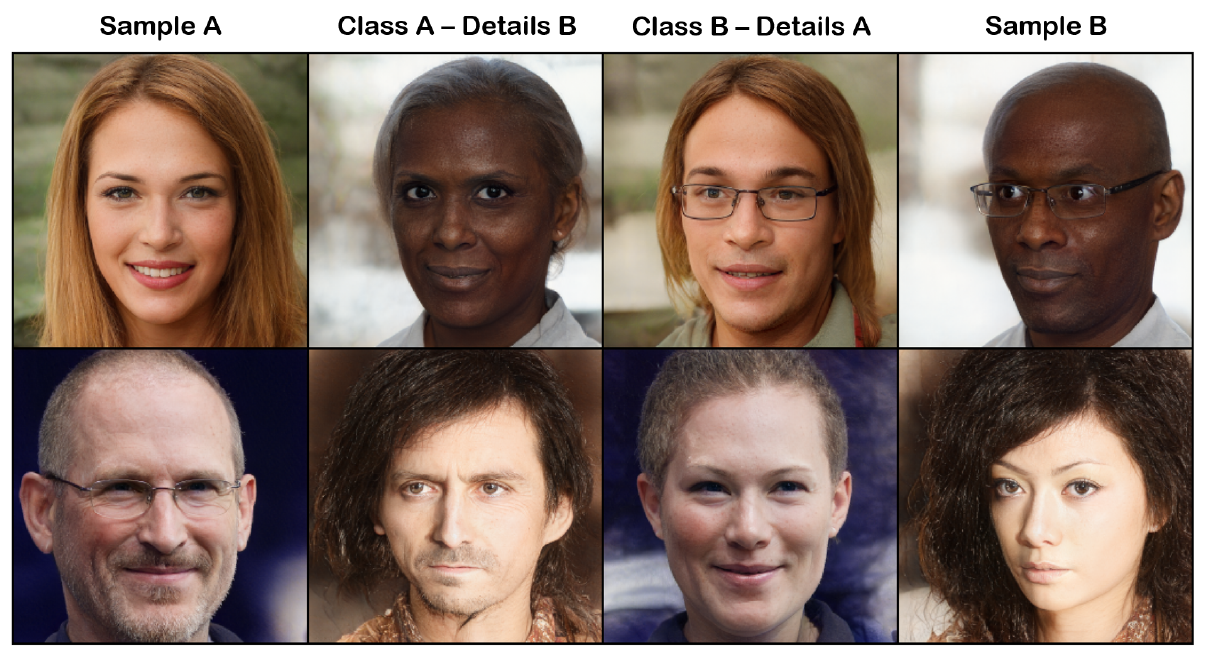}
  \caption{Mixing samples of different classes following what has been learned by Bayesian Optimization on the Celeb-A HQ Gender task. From left to right: Sample A (female or male), Class features of sample A mixed with Details of sample B, Class features of sample B mixed with Details of sample A, Sample B (male or female).}
  \label{fig:mixing_gender}
\end{figure*}

\begin{figure*}
  \centering
  \includegraphics[width=1.0\linewidth]{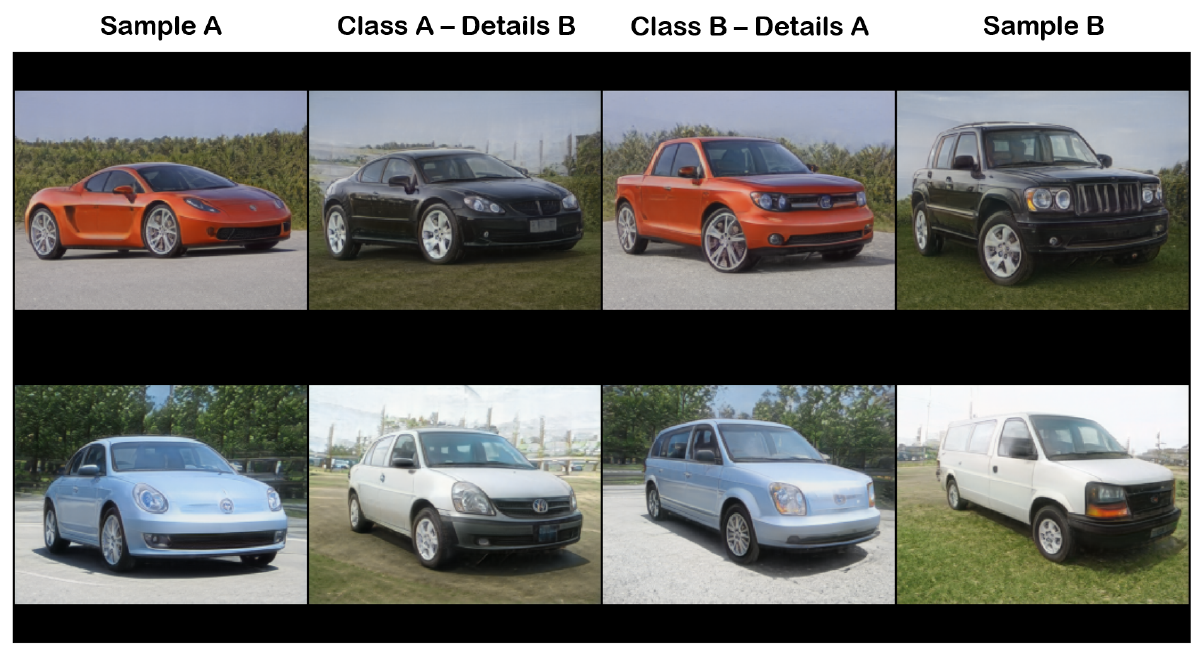}
  \caption{Mixing samples of different classes following what has been learned by Bayesian Optimization on the Stanford Cars Type Classification task. From left to right: Sample A (coupe or hatchback), Class features of sample A mixed with Details of sample B, Class features of sample B mixed with Details of sample A, Sample B (suv or van).}
  \label{fig:mixing_cars}
\end{figure*}

In \Cref{fig:gender_combinations,fig:ids_combinations,fig:cars_combinations} (b) of the main paper we show the hyperparameter values \textbf{learned} by Bayesian Optimization, compared to the \textbf{linear} and \textbf{cosine} case. For the identity classification task, the combination is monotonic and matches the idea that MLVGMs represent features in global-to-local manner, with class relevant information mainly contained in the first latent levels, and gradually decreasing. However, for the coarse-grained tasks (gender and car types classification), the \textbf{learned} combinations tend to assign low $\alpha$ values to specific intermediate latents. These are $i=2, 5, 6$ and $i=1, 2, 4, 5, 9, 13$ for the two tasks, respectively. In other terms, Bayesian Optimization suggests to maintain the original information of these latent levels, and to discard and re-sample the remaining codes. This means that the optimization process defines two types of codes: ``class codes'' which should maintain original information, and ``detail codes'' that can be re-sampled. In \Cref{fig:mixing_gender,fig:mixing_cars} we qualitatively verify this aspect, by mixing the class and detail codes given by BO of two samples (A and B), of \emph{different} classes. The second and third column in the figures shows what happens when mixing class codes of one sample with detail codes of the other. As visible, 
it is true that class-relevant and irrelevant information is disentangled, allowing to change the details of an image without altering its label. This qualitative analysis suggests that Bayesian Optimization is effective in distinguishing class-relevant features, assigning low $\alpha$ values to the corresponding latent levels. Furthermore, the visualization supports the hypothesis made in the main paper: in coarse-grained classification tasks, the class-relevant information is contained in a few, intermediate latents.

\section{Experimental Details}
\label{supp:exp_details}

\paragraph{Obtaining the pre-trained MLVGMs.}

For the Celeb-A HQ and Cars tasks we use the official StyleGan-2 checkpoints, coupled with the corresponding encoders. The public repositories are \href{https://github.com/omertov/encoder4editing}{https://github.com/omertov/encoder4editing} for faces and \href{https://github.com/sapphire497/style-transformer}{https://github.com/sapphire497/style-transformer} for cars. These allow to download the pre-trained Autoencoders, based on StyleGan-2.\\
Regarding NVAE, we train our own model on the Celeb-A dataset at resolution $64\times64$. The model has $3$ scales of $8$ groups, for a total of $24$ latent variables. We trained the model for $600$ epochs, with a cumulative batch size of $256$. The learning rate has been decayed from $1e-3$ to $5e-4$. The total number of parameters is $705.678$ M. The pre-trained model, code and complete training configuration are available at \href{https://github.com/SerezD/NVAE-from-scratch}{https://github.com/SerezD/NVAE-from-scratch}.

\paragraph{Creating the Datasets.}

The instructions on how to download the gender classification dataset are available at: \href{https://github.com/ndb796/CelebA-HQ-Face-Identity-and-Attributes-Recognition-PyTorch}{https://github.com/ndb796/CelebA-HQ-Face-Identity-and-Attributes-Recognition-PyTorch}. For Celeba-64 identity classification, we obtain the original data and annotated identities from the Celeb-A project page: \href{https://mmlab.ie.cuhk.edu.hk/projects/CelebA.html}{https://mmlab.ie.cuhk.edu.hk/projects/CelebA.html}. The dataset comprises $10.177$ identities. First, we center crop and resize each image to achieve the correct resolution of $64 \times 64$. Then, we select a random subset of $100$ identities, between these that have at least $15$ samples per class. For Stanford Cars, we download the original dataset from Kaggle, at: \href{https://www.kaggle.com/datasets/jessicali9530/stanford-cars-dataset}{https://www.kaggle.com/datasets/jessicali9530/stanford-cars-dataset}. We use the provided class names to filter those containing the words ``coupe'', ``hatchback'', ``suv'' or ``van'' in the original label. Then, we regroup them into $4$ classes and manually remove outliers, keeping the same number of images per class. Lastly, each image is squared, adding black padding at the top and bottom sides, and resized at $128 \times 128$ resolution.

\paragraph{Training of classifiers.}
We use \verb|pytorch.models| classes for the base architecture of each classifier. These are: \verb|resnet50|, \verb|vgg11_bn|, and \verb|resnext50_32x4d|, respectively. In all cases, we replace the final fully connected layer with an MLP composed of two linear layers, a batch normalization, and a ReLU activation function. Then, we train the full model (backbone + head) on each task.

\paragraph{Hyperparameters of attacks.}
In a preliminary phase, we tested several hyperparameter configurations for both attacks, ensuring a sufficient number of steps for convergence. In general, we observe better performance of DeepFool, requiring significantly fewer steps to achieve high success rates. We show the final values for each task in \Cref{tab:deepfool,tab:cw}.

\begin{table}
  \centering
  \begin{tabular}{@{}lccc@{}}
    \toprule
    Task & Class Tested & Overshoot & Steps \\
    \midrule
    Gender     & $2$ & $0.01$ & $1024$ \\
    Identities & $8$ & $0.02$ & $128$ \\
    Cars       & $4$ & $0.02$ & $256$ \\
    \bottomrule
  \end{tabular}
  \caption{Final hyperparameters for DeepFool attack.}
  \label{tab:deepfool}
\end{table}

\begin{table}
  \centering
  \begin{tabular}{@{}lccccc@{}}
    \toprule
    Task & Reg. Const. & Conf. & Steps & Restarts & LR\\
    \midrule
    Gender     & $64$ & $0.01$ & $1024$ & $8$ & $1e-3$\\
    Identities & $16$ & $0.05$ & $1024$ & $8$ & $5e-3$\\
    Cars       & $24$ & $0.02$ & $1024$ & $8$ & $2e-3$\\
    \bottomrule
  \end{tabular}
  \caption{Final hyperparameters for Carlini \& Wagner attack.}
  \label{tab:cw}
\end{table}

\paragraph{Code and Training of competitors.}

We use the official code provided by each competitor to train models on our tasks. The code is available at \href{https://github.com/yaodongyu/TRADES}{https://github.com/yaodongyu/TRADES} for TRADES, at \href{https://github.com/nercms-mmap/A-VAE}{https://github.com/nercms-mmap/A-VAE} for A-VAE , and at \href{https://github.com/shayan223/ND-VAE}{https://github.com/shayan223/ND-VAE} for ND-VAE. For TRADES, we fine-tune each classifier for $50$ additional epochs, setting the beta regularization term to $1.5, 1.0, 8.0$ for gender, identities, and cars, respectively. We found these values to give the best trade-off between robustness and accuracy. For A-VAE, we follow the indications of the original paper and downsample the initial image to $32 \times 32$, independently from the starting resolution. We continue training until convergence of the GAN model. For ND-VAE, we generate the adversarial training set with FGSM attack. The autoencoder model has $2$ scales with $2$ groups for the gender and cars tasks, and $1$ scale only in the identity case, due to lower starting resolution. We train each model for $100$ epochs.

\section{Ablation Studies}
\label{supp:ablations}

\begin{figure}
  \centering
  \includegraphics[width=1.0\linewidth]{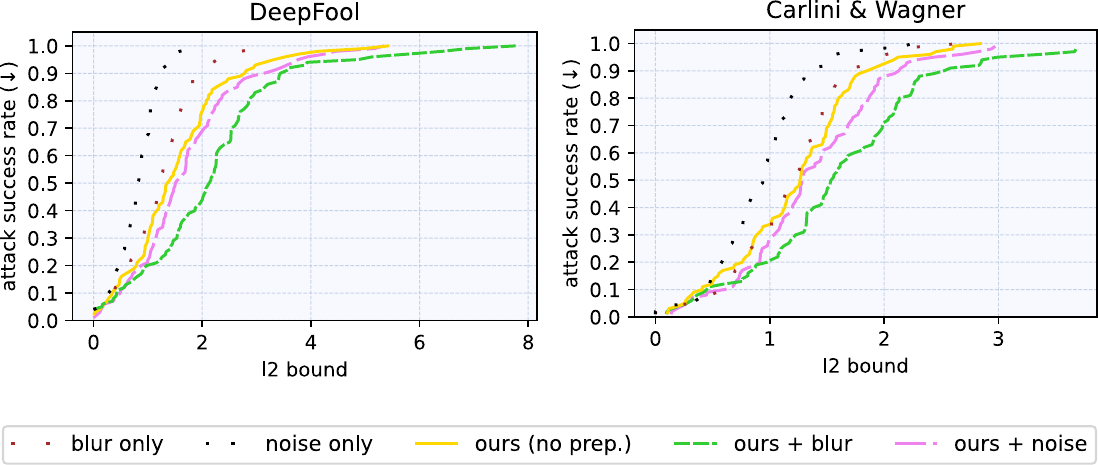}
  \caption{Attack succes rate (the lower the better) for increasing $\textit{L}_2$ bounds for different attacks and preprocessing operations, on the Celeb-A HQ gender task. Dotted lines show the lower bound of applying only blur or noise as a defense mechanism. }
  \label{fig:ablations_gender}
\end{figure}
\begin{figure}
  \centering
  \includegraphics[width=1.0\linewidth]{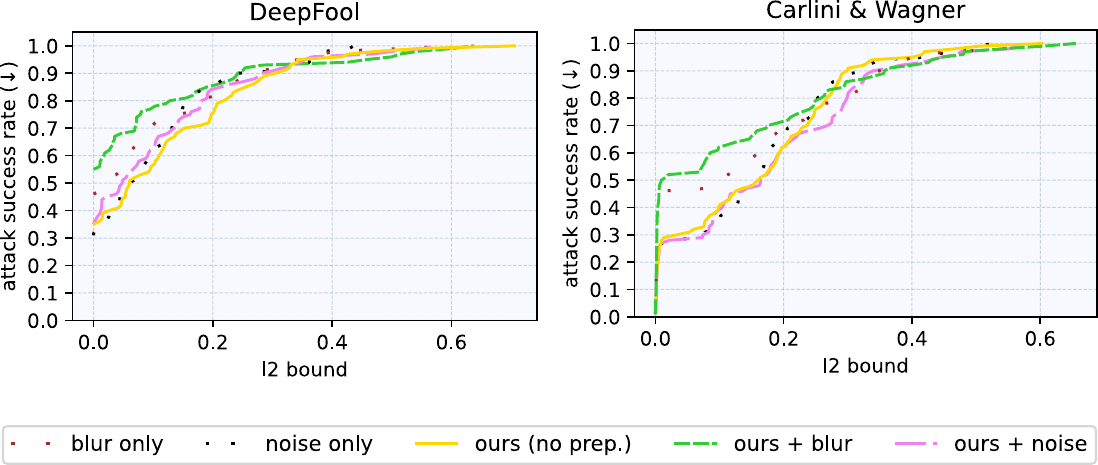}
  \caption{Attack succes rate (the lower the better) for increasing $\textit{L}_2$ bounds for different attacks and preprocessing operations, on the Celeb-A 64 identities task. Dotted lines show the lower bound of applying only blur or noise as a defense mechanism. }
  \label{fig:ablations_ids}
\end{figure}
\begin{figure}
  \centering
  \includegraphics[width=1.0\linewidth]{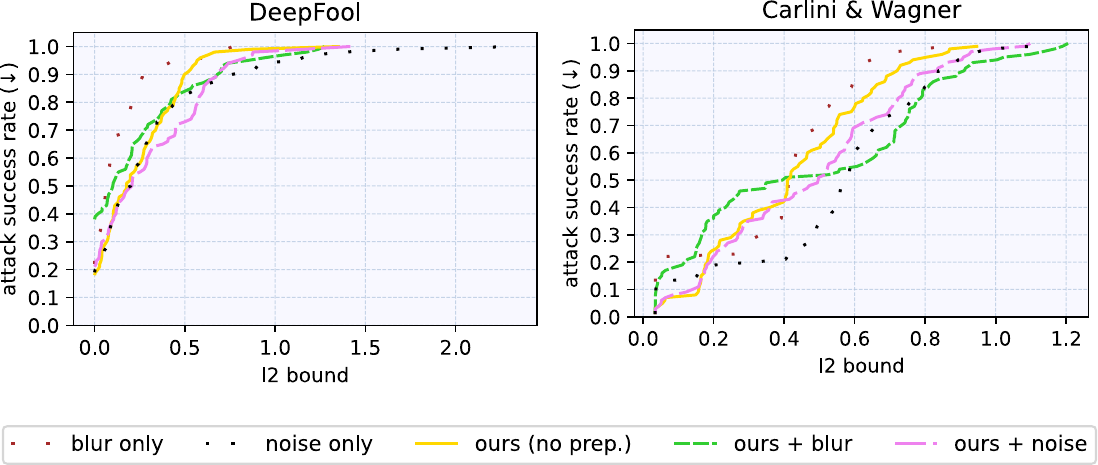}
  \caption{Attack succes rate (the lower the better) for increasing $\textit{L}_2$ bounds for different attacks and preprocessing operations, on the Stanford Cars 128 task. Dotted lines show the lower bound of applying only blur or noise as a defense mechanism. }
  \label{fig:ablations_cars}
\end{figure}
In \Cref{fig:results_gender,fig:results_ids,fig:results_cars} (c) of the main paper we test what type of $\alpha$ combination (\textbf{learned}, \textbf{linear} or \textbf{cosine}) performs best on each task. Here, we test the effect of preprocessing operations (adding gaussian noise or applying gaussian blur) when applied on the best configuration. For Gaussian noise, we add a random perturbation $\nu$ to each image, where $|\nu|_2 = 4$ in the gender and cars tasks, while $|\nu|_2 = 2$ on the identities task. For Gaussian blur we set the kernel size to $2^{\frac{R}{2}} - 1$, where $R$ is the image resolution, and keep $\sigma = 1$ everywhere. Results for each task are visible in \Cref{fig:ablations_gender,fig:ablations_ids,fig:ablations_cars}. For better comparison, the attack success rate is computed also using Gaussian Noise or Gaussian Blur as a standalone purification mechanism. In the gender case, the combination with Blur is beneficial to the method, while adding Gaussian noise performs worse. On the remaininig tasks, the addition of Gaussian Noise allows an extra boost on the Carlini \& Wagner attack, while all runs achieve similar performances on DeepFool. For computing the final results, shown in \Cref{fig:results_gender,fig:results_ids,fig:results_cars} of the main paper, we use the best $\alpha$ + preprocessing combination found after these ablation studies.

\section{Qualitative Results}
\label{supp:qualitative_examples}
\begin{figure*}
  \centering
  \includegraphics[width=1.0\linewidth]{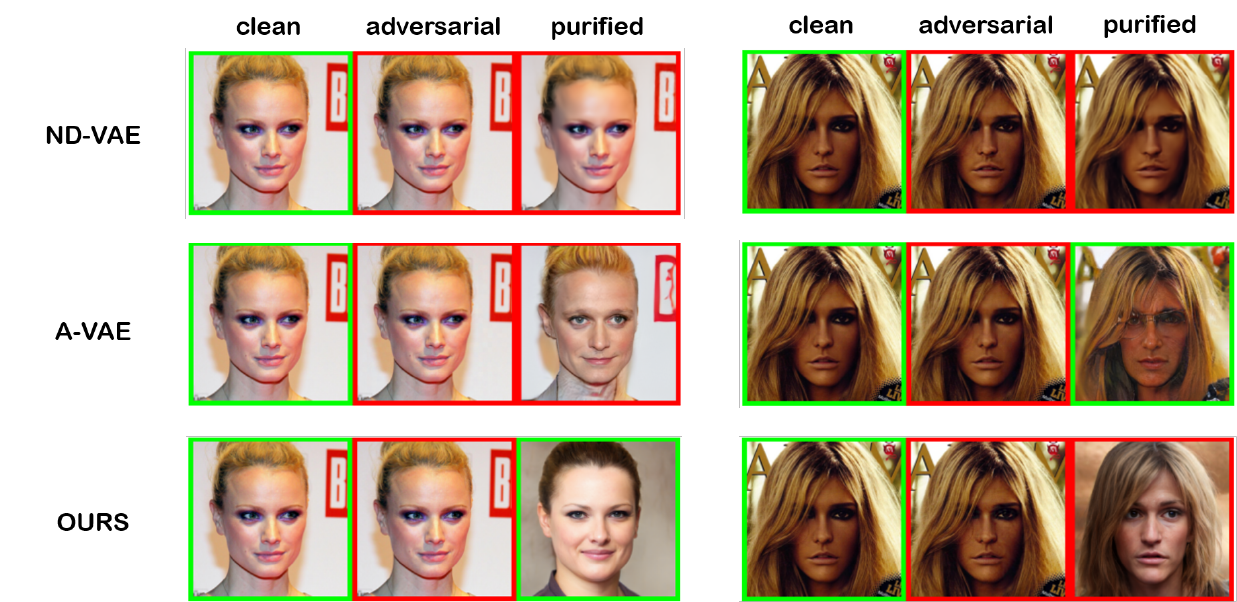}
  \caption{Qualitative examples of purification via MLVGMs on the Celeb-A HQ gender task. We show one success case (left) and failure case (Right), comparing the purification with the one of ND-VAE and A-VAE. In each example, we additionaly show the clean and adversarial images.}
  \label{fig:qualitative_gender}
\end{figure*}
\begin{figure*}
  \centering
  \includegraphics[width=1.0\linewidth]{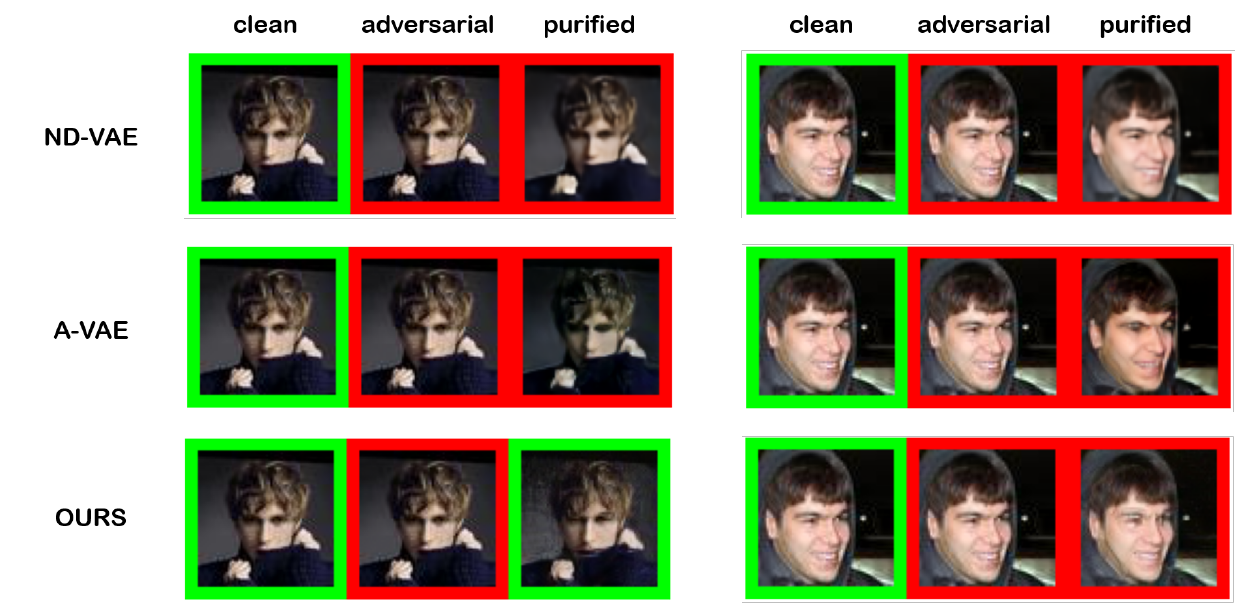}
  \caption{Qualitative examples of purification via MLVGMs on the Celeb-A 64 identities task. We show one success case (left) and failure case (Right), comparing the purification with the one of ND-VAE and A-VAE. In each example, we additionaly show the clean and adversarial images.}
  \label{fig:qualitative_ids}
\end{figure*}
\begin{figure*}
  \centering
  \includegraphics[width=1.0\linewidth]{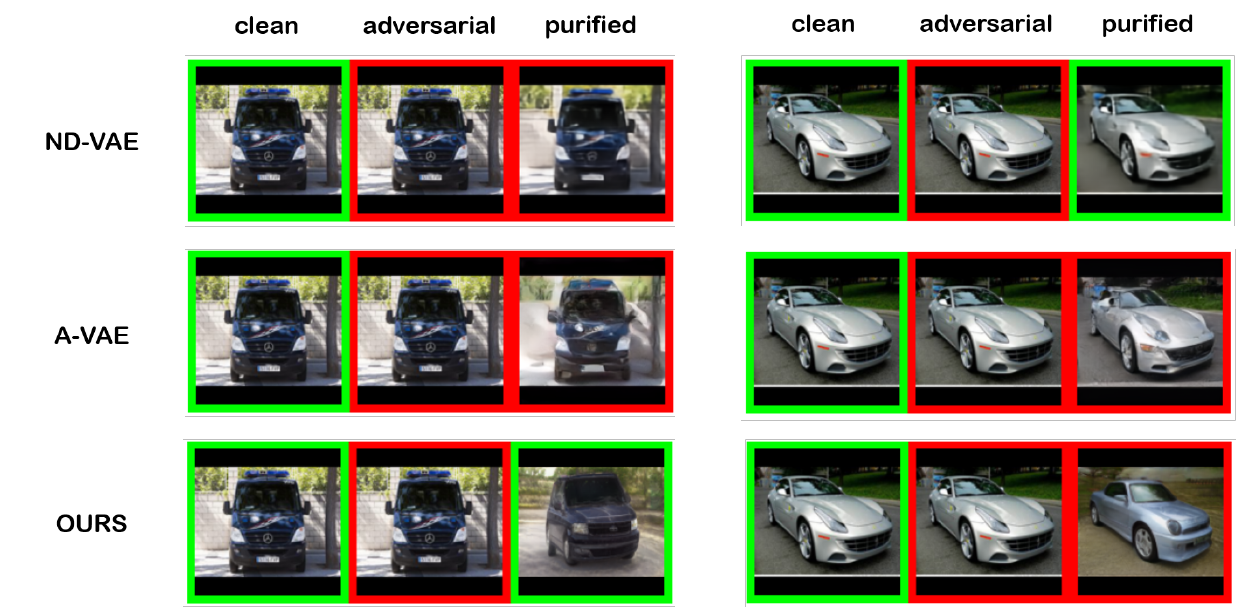}
  \caption{Qualitative examples of purification via MLVGMs on the Stanford Cars 128 task. We show one success case (left) and failure case (Right), comparing the purification with the one of ND-VAE and A-VAE. In each example, we additionaly show the clean and adversarial images.}
  \label{fig:qualitative_cars}
\end{figure*}
We show one success and one failure case of purification on each task in \Cref{fig:qualitative_gender,fig:qualitative_ids,fig:qualitative_cars}, respectively. For additional comparison, the images also show the same examples when purified with ND-VAE and A-VAE methods. As visible, the former mainly acts as a denoiser, attempting to remove adversarial noise while reconstructing the clean sample. A-VAE tries instead to modify the details of the input sample, while maintaining the semantic content. However, the mechanism appears limited, and the final purified image highly resambles the initial image. On the opposite, our method based on MLVGMs aims at maintaining only the relevan class information, while re-sampling all remaining irrelevant details. In coarse-grained classification (like male/female, with only two classes), this means the the final purified samples are highly different from the initial image. Conversely, in fine-grained tasks such as identity classification, the combination of $\alpha$ hyperparameters is properly tuned to alter only a few details, since it is easier to alter the class label. With this approach, MLVGMs can be effective both in coarse and fine-grained classification tasks, while limiting the freedom of the attacker, which is forced to act on the initial image by changing only imperceptible details.

\section{Additional Results on Autoattack}
\label{supp:autoattack}
\begin{figure*}
  \centering
  \includegraphics[width=1.0\linewidth]{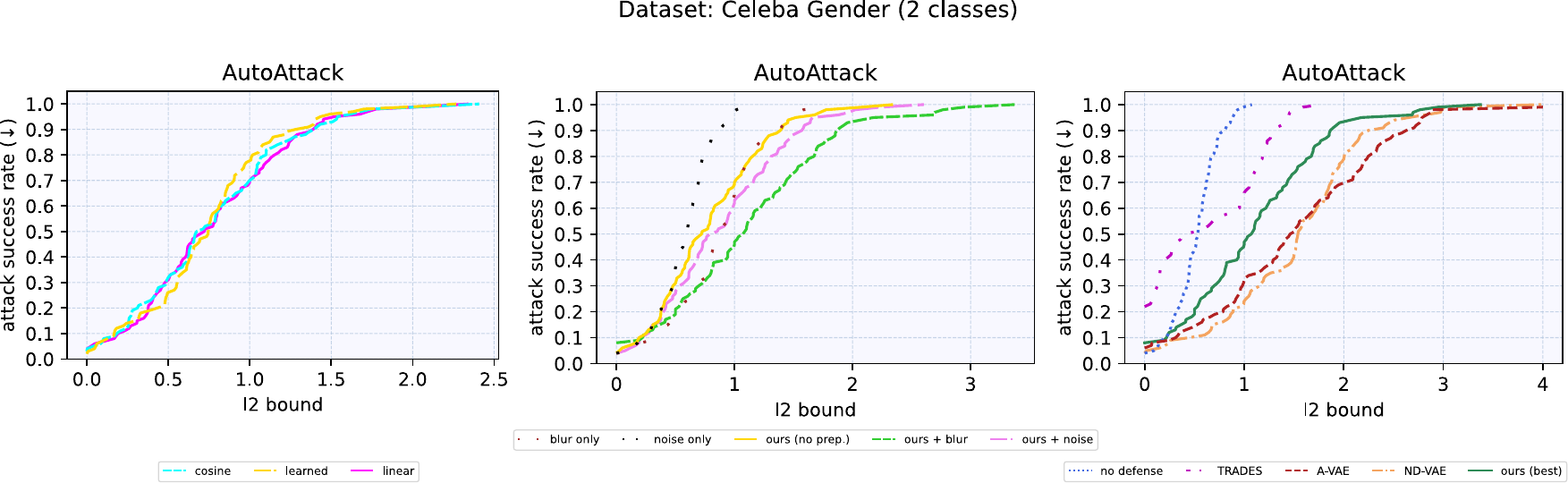}
  \caption{Success rates of Autoattack \cite{autoattack} for $100$ samples on the Celeb-A HQ gender classification task. From left to right: comparison of the tested combinations (\textbf{learned}, \textbf{linear} and \textbf{cosine}); ablations on the introduction of random noise and blur; and comparison with other methods.}
  \label{fig:autoattack_gender}
\end{figure*}
\begin{figure*}
  \centering
  \includegraphics[width=1.0\linewidth]{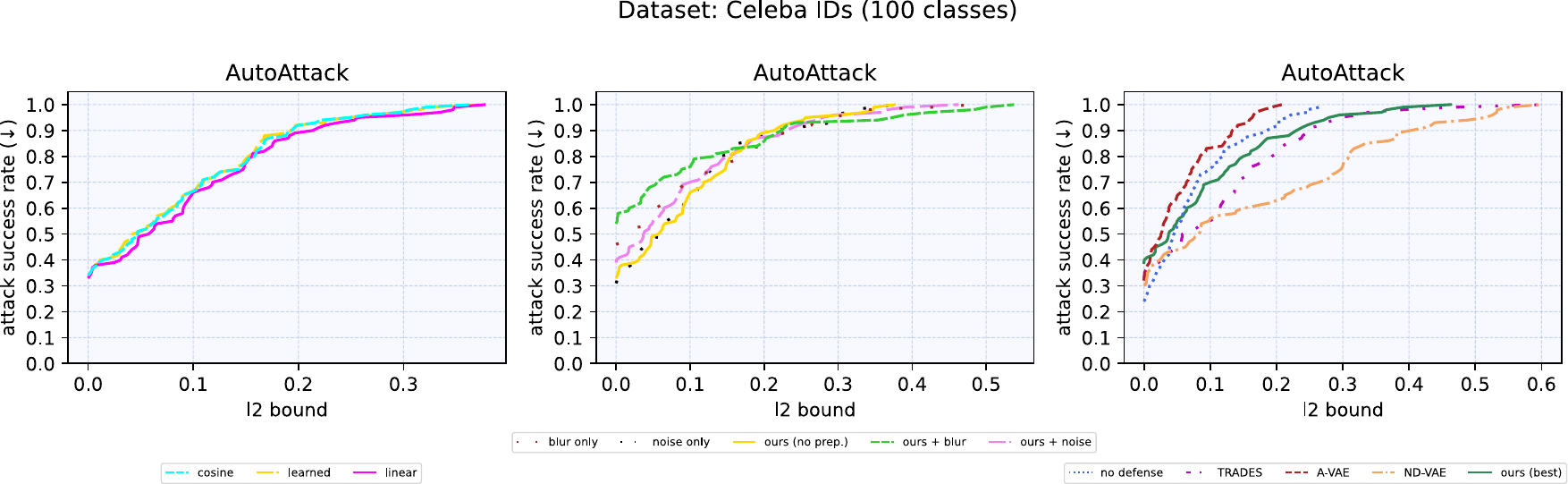}
  \caption{Success rates of Autoattack \cite{autoattack} for $100$ samples on the Celeb-A ids classification task. From left to right: comparison of the tested combinations (\textbf{learned}, \textbf{linear} and \textbf{cosine}); ablations on the introduction of random noise and blur; and comparison with other methods.}
  \label{fig:autoattack_ids}
\end{figure*}
\begin{figure*}
  \centering
  \includegraphics[width=1.0\linewidth]{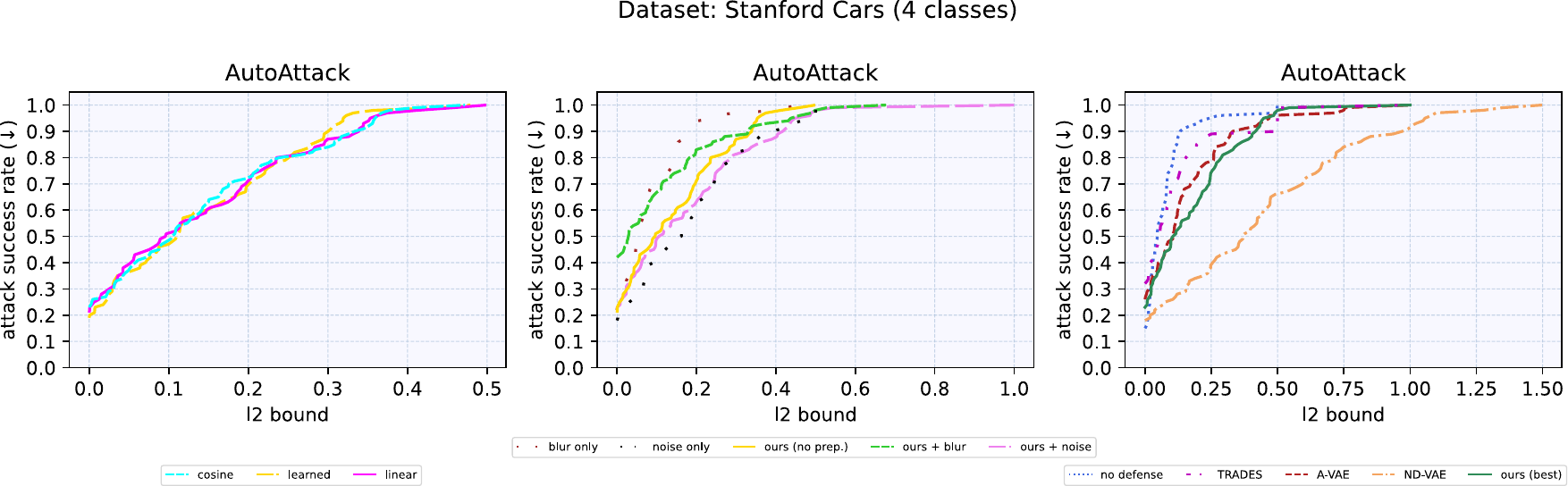}
  \caption{Success rates of Autoattack \cite{autoattack} for $100$ samples on the Stanford Cars classification task. From left to right: comparison of the tested combinations (\textbf{learned}, \textbf{linear} and \textbf{cosine}); ablations on the introduction of random noise and blur; and comparison with other methods.}
  \label{fig:autoattack_cars}
\end{figure*}
In completion of the results reported in \Cref{sec:experiments}, we analyze the performance of our method also with Autoattack \cite{autoattack}, a well-known evaluation pipeline which attempts to fool defense mechanisms by combining different techniques. The results for the three tasks are shown in \Cref{fig:autoattack_gender,fig:autoattack_ids,fig:autoattack_cars}. In general, changing the hyperparameters combination (\textbf{learned}, \textbf{linear} or \textbf{cosine}) does not seem to have a significant effect in this scenario, while adding gaussian blur or noise is beneficial in Celeb-A HQ and Standford Cars tasks, respectively. When comparing with other methods, the general trend observed in \Cref{sec:experiments} is maintained, with ND-VAE performing best on ids and cars classification and our MLVGMs-based mechanism achieving comparable results with other methods. Interestingly, TRADES \cite{trades} fails against this attack in the Celeb-A HQ gender classification task, showing the effectiveness of AutoAttack even when compared to state-of-the-art methods such as C\&W or DeepFool. 

\end{document}